\documentclass[12pt,a4paper]{article}

\usepackage[T1]{fontenc}
\usepackage[utf8]{inputenc}
\usepackage[margin=2.5cm, top=3.5cm]{geometry}
\usepackage{setspace}
\usepackage{hyperref}
\usepackage{url}
\usepackage{natbib}
\usepackage{titlesec}
\usepackage{enumitem}
\usepackage{xcolor}
\usepackage{csquotes}
\usepackage{longtable} 

\usepackage{helvet}        

\usepackage{fancyhdr}
\hypersetup{
  colorlinks=true,
  linkcolor=black,
  citecolor=black,
  urlcolor=teal,
  pdftitle={AI-Driven Science: Promises, Risks, and Open Questions},
  pdfauthor={Emmanuel Jeannot}
}

\titleformat{\section}{\large\bfseries}{\thesection.}{0.5em}{}[\titlerule]
\titleformat{\subsection}{\normalsize\bfseries}{\thesubsection.}{0.5em}{}

\title{%
  \textbf{The Industrialization of Research}\\[0.5em]
  {\large\itshape On AI-Driven Science and Its Consequences}%
}

\author{Emmanuel Jeannot\\[0.5em]
  \normalsize Inria\\[0.3em]
  \normalsize \href{mailto:emmanuel.jeannot@inria.fr}{emmanuel.jeannot@inria.fr}}
\date{June 2026 \\ \small Version 3}

\begin{document}

\maketitle
\thispagestyle{empty}

\begin{abstract}
\noindent
Artificial intelligence is transforming scientific research — not merely as a more 
powerful instrument, but as an autonomous participant in the research cycle itself. 
This transition constitutes, in the most precise sense of the term, the 
industrialization of research: a shift from a craft model, in which knowledge, method, 
and judgment are embedded in the researcher, to a pipeline model, in which these steps 
are decomposed, automated, and supervised. The US Department of Energy's Genesis 
Mission is the most ambitious current instantiation of this shift, but the fundamental
questions it raises extend far beyond any single program. This essay examines seven 
such questions: the erosion of the intergenerational transmission of scientific 
competence; the growing opacity of AI-generated theories; the collapse of peer 
evaluation under a flood of machine-generated output; the unproven capacity of AI for 
paradigm-shifting discovery; the capture of the scientific agenda by political and 
industrial actors; the compounding of systematic errors in closed-loop pipelines; and 
the structural bifurcation of the global research community into incommensurable tiers. 
These concerns do not constitute an argument against AI-driven science — whose 
demonstrated potential is real and significant. They constitute the conditions under 
which that potential can be responsibly pursued.

\end{abstract}
\newpage
\tableofcontents
\newpage

\section*{Introduction}
\addcontentsline{toc}{section}{Introduction}

Artificial intelligence is a genuine game changer. Across medicine, engineering, 
education, and creative work, it is disrupting established practices at a pace and 
scale that few technologies in history have matched. As Pope Leo~XIV's encyclical 
\textit{Magnifica Humanitas}~\cite{magnifica_humanitas_2026} reminds us, this 
disruption is not inevitable in its consequences — it is a tool whose character depends 
entirely on the ends it serves and the conditions under which it is deployed.

Using AI to accelerate scientific discovery is, we believe, one of the most 
consequential and promising directions available to us. AlphaFold~\cite{alphafold2021} 
transformed molecular biology in a matter of months. ML models have identified new 
antibiotics \cite{antibiotics2023,antibiotics2024}, predicted material properties 
\cite{gnome2023}, and detected patterns in astronomical datasets that would have taken 
human researchers decades to find. LLMs have independently solved long-standing open 
problems~\cite{knuth_claude_2026} and disproved decades-old 
conjectures~\cite{openai_unit_distance_2026}. This is not hype — it is a demonstrated 
and expanding capability.

By \emph{AI-driven science}, we mean more specifically the deployment of AI systems 
— in particular large language models and autonomous agents — as active participants in 
the scientific research cycle: formulating hypotheses, designing experiments, analyzing 
results, and proposing next steps, with varying degrees of human supervision. This is 
distinct from AI as a computational instrument, where the system executes a well-defined 
task within boundaries set by human researchers. The difference is not one of degree 
but of kind: in the first case, AI accelerates what humans do; in the second, AI begins 
to do what humans do. Pursued at scale, this transition amounts to what we call the 
\emph{industrialization of research} — a shift from a craft model of scientific 
production, in which knowledge and judgment are embedded in the researcher, to a 
pipeline model, in which these steps are decomposed, automated, and supervised. 

The industrialization of research, as AI-driven science proposes it, is not without 
promise — but it raises systemic questions that must be addressed in parallel with 
the deployment, not after it. Large-scale initiatives such as the Genesis Mission, 
which aims to double scientific productivity within a decade by placing AI agents at 
the center of the full research cycle, make these questions urgent. This essay 
addresses them, in the spirit of the Leiden Declaration on Artificial Intelligence and 
Mathematics \cite{leiden_declaration_2026}, focusing on the consequences for 
scientific practice. 

\section{The Industrialization of Research}
\label{sec:industrialization}
What AI-driven science proposes is, in the most precise sense of the term, the 
\emph{industrialization of research}. The analogy with the industrial revolutions of 
the nineteenth and twentieth centuries is not rhetorical — it is structural\footnote{This
parallel is not unique to this essay. Sylvia Serfaty, in her recent book 
\textit{Des équations personnelles}~\cite{serfaty_equations_2026}, uses precisely 
the same image: mathematics, she writes, has been an ``artisanal'' practice for more 
than two thousand years, and mathematicians must now climb into a ``racing car'' to 
face what amounts to an ``industrial revolution''.}. In both 
cases, a craft model of production gives way to a pipeline model: tasks that were 
previously integrated in a single practitioner or a small team are decomposed, standardized, and 
delegated to machines, while the human role shifts from execution to supervision. The 
craftsman who mastered the full arc of a trade — material, tool, judgment, and 
result — is replaced by the operator who monitors a system whose inner workings 
exceed any single person's grasp.

In the craft model of science, a researcher identifies a question, designs an 
approach, executes experiments, interprets results, draws conclusions and present
the work among its peers and to the society, exercising 
judgment at every step. Knowledge, method, and intuition are embedded in the person 
and transmitted through apprenticeship. In the pipeline model that AI-driven science 
proposes, these steps are decomposed and partially automated: agents handle hypothesis 
generation, experimental design, data collection, analysis, and validation, while 
human researchers supervise, validate, and orient the system. The result is a 
qualitative change in the relationship between the researcher and the scientific 
process — and, potentially, in the nature of the knowledge produced.

The parallel with industrial history is very instructive. The industrial revolution 
produced real and substantial benefits — gains in productivity, speed, and scale that 
no craft system could have matched — alongside real and substantial costs that were 
systematically ignored at the moment of deployment: environmental externalities, 
concentration of capital, loss of craft knowledge, and asymmetric access between those 
who owned the infrastructure and those who supplied it with labor. The same structure 
of benefits and costs appears to be reproducing itself in the industrialization of 
research, as Table~\ref{tab:industrial} illustrates. What makes the current transition 
particularly significant is not that it breaks the parallel, but that it deepens it: 
the industrial revolution automated \emph{physical labor} while leaving intellectual labor 
intact; the industrialization of research extends that same logic to the \emph{mind itself}. 
Science — one of the most distinctly human of cognitive practices — becomes the next domain 
to be organized around machines, workflows, and supervisors rather than craftsmen.


\begin{longtable}{|p{3.5cm}|p{4cm}|p{4cm}|p{1cm}|}
\hline
\textbf{Generic dimension} & \textbf{Industrial Revolution} & 
\textbf{Research Industrialization} & \textbf{Sec.} \\
\hline
\multicolumn{4}{|l|}{\textit{Advantages}} \\
\hline
Output per practitioner & Massive increase in goods per worker & 
Massive increase in results per researcher & 
\S\ref{sec:why} \\
\hline
Cycle time reduction & Weeks $\to$ hours for manufactured goods & 
Years $\to$ days for hypothesis-to-result cycles & 
\S\ref{sec:why} \\
\hline
Reach beyond manual limits & Production volumes unreachable by hand & 
Hypothesis spaces unreachable by human teams & 
\S\ref{sec:why} \\
\hline
Cross-domain recombination & Cross-industry innovation (steel + railways) & 
LLMs trained on all disciplines simultaneously & 
\S\ref{sec:why} \\
\hline
Reduction of individual variance & Standardization reduces worker variation & 
No confirmation bias, no disciplinary blind spots & 
\S\ref{sec:why} \\
\hline
Broadening of access & Goods became affordable to more people & 
Non-specialists can produce scientific results & 
\S\ref{sec:why} \\
\hline
Unplanned spillovers & New technologies (materials, transport) & 
Interdisciplinary discoveries at scale — yet to be demonstrated & 
\S\ref{sec:why} \\
\hline
\multicolumn{4}{|l|}{\textit{Neutral / Ambivalent}} \\
\hline
Role shift for the practitioner & Craftsman $\to$ machine operator & 
Scientist $\to$ AI supervisor & 
\S\ref{sec:competence} \\
\hline
Capacity for radical innovation & New industries created new paradigms & 
Whether AI can produce paradigm shifts remains unproven & 
\S\ref{sec:paradigm} \\
\hline
Uniformity of output & Products became uniform — variety reduced & 
Science may become uniform — diversity of approaches reduced & 
\S\ref{sec:capture} \\
\hline
\multicolumn{4}{|l|}{\textit{Risks and costs}} \\
\hline
Loss of tacit knowledge transmission & Apprenticeship replaced by on-the-job training & 
Mentorship cycle student $\to$ researcher $\to$ mentor disrupted & 
\S\ref{sec:competence} \\
\hline
Loss of prior capabilities & Craft skills lost in one generation & 
Scientific intuition and training pipeline at risk & 
\S\ref{sec:competence} \\
\hline
Loss of end-to-end comprehension & Workers lost understanding of the full production process & 
Researchers may not understand the results AI produces & 
\S\ref{sec:comprehension} \\
\hline
Concentration of ownership & Factories owned by few & 
AI infrastructure owned by few actors & 
\S\ref{sec:capture} \\
\hline
Capture of production priorities & Production driven by owners and market & 
Research driven by political and industrial priorities & 
\S\ref{sec:capture} \\
\hline
Energy and environmental cost & Coal and steam — externality long ignored & 
Data centers — energy externality currently underweighted & 
\S\ref{sec:capture} \\
\hline
Asymmetry of access & Rich nations industrialized first; gap persisted & 
Well-funded institutions first — gap may be permanent & 
\S\ref{sec:bifurcation} \\
\hline
Extraction from the periphery & Colonies supplied raw materials to industry & 
Less-resourced institutions supply results to systems they cannot access & 
\S\ref{sec:bifurcation} \\
\hline
\caption{The industrialization of research: a historical parallel. 
Section references indicate where each dimension is examined.}
\label{tab:industrial}
\end{longtable}

Table~\ref{tab:industrial} draws the parallel with industrial history, but not all 
dimensions are treated equally in this essay, which focuses specifically on the 
impact of AI-driven science on how research is conducted, transmitted, and evaluated.
The advantages of AI-driven science and 
the constitutive reasons that motivate its development are examined in 
Section~\ref{sec:why}. The more complex consequences — for competence transmission, 
scientific comprehension, paradigm shifts, agenda capture, and access equity — are 
addressed in Sections~\ref{sec:competence} to~\ref{sec:bifurcation}. Two dimensions 
specific to AI-driven science that have no direct industrial precedent receive dedicated 
treatment: the collapse of research evaluation under AI-generated volume 
(Section~\ref{sec:evaluation}), and the compounding of errors in closed-loop pipelines 
(Section~\ref{sec:errors}). The broader societal and environmental impacts of 
large-scale AI deployment — energy costs, democratic governance, concentration of 
economic power — are noted where they bear directly on scientific practice but deserve 
separate treatment beyond the scope of this essay.
\section{Why AI-Driven Science?}
\label{sec:why}

Before examining the fundamental questions that large-scale AI-driven science raises, it is worth asking a prior 
question: why pursue it in the first place? The answer is not self-evident, and it matters — because the strength of 
the case for AI-driven science is precisely what makes the concerns that follow worth taking seriously.

The term ``\emph{AI for science}'' covers two qualitatively different things, and conflating them creates confusion.

The first is \textbf{AI as an algorithmic tool} — a system designed and trained by humans to solve a specific, 
well-defined class of problems. AlphaFold predicts protein structures from amino acid sequences \cite{alphafold2021}. 
Graph neural networks reconstruct particle collisions at the LHC \cite{cms_mlpf_2026}. GNoME identifies stable crystal 
candidates from materials databases \cite{gnome2023}. In each case, the architecture, the training objective, and the 
problem framing are human choices; the AI executes within those boundaries with extraordinary efficiency. This is a 
authentic and important contribution — but it is, in nature, an advanced instrument, not an independent scientific actor.

The second is \textbf{AI as a (semi-)autonomous research assistant} — a general-purpose system given an open problem, 
which explores, reformulates, and arrives at a solution through its own iterative reasoning, with minimal human 
guidance. Two recent results illustrate this mode. In February 2026, Donald Knuth described how Claude Opus 4.6 solved 
in a few hours and some thirty iterative explorations a combinatorics problem he had been working on for several weeks
\cite{knuth_claude_2026}. In May 2026, an internal 
OpenAI model disproved an 80-year-old conjecture by Erd\H{o}s on unit distances in the plane — one of the best-known 
open problems in combinatorial geometry — by producing a family of counterexamples using sophisticated tools from 
algebraic number theory that no one had previously thought to apply.

This essay is concerned primarily with the second mode, and with what happens when it 
is deployed at the scale of national and international science infrastructure. The 
motivations are real and deserve to be stated clearly. As shown above, difficult 
scientific results can already be achieved semi-autonomously by LLMs — and this is no 
accident. These results follow from intrinsic properties that make LLMs particularly 
well-suited to scientific research:
they are trained on a corpus that spans the entirety of human 
scientific knowledge and the full conceptual vocabulary of all disciplines 
simultaneously, making them natural instruments for interdisciplinary research because 
the connections between fields are already embedded in their representations. Beyond 
interdisciplinarity, AI systems can explore hypothesis spaces in parallel at speeds no 
team can match; they can read and ingest new results efficiently and rapidly; they are 
free from the confirmation biases that lead entire communities to miss promising 
directions; they can accelerate the full cycle from hypothesis to result in ways that 
compound across iterations; and their capacity for parallel deployment allows a scale 
of scientific production that no human workforce could sustain.

These are serious advantages. They justify serious investment, and they explain why initiatives like the Genesis 
Mission are being pursued with urgency. The question is not whether these advantages are real. The question is what 
else comes with them.

\section{The Genesis Mission: A Case Study in AI-Driven Science}
\label{sec:genesis}

Announced by the United States Department of Energy in late 2025, the Genesis Mission is one of the most ambitious 
AI-driven scientific initiatives to date — a national effort to build an integrated AI-for-science infrastructure 
connecting supercomputers, AI systems, and quantum technologies across the DOE national laboratories 
\cite{genesis_eo_2025}. Rick Stevens, Associate Laboratory Director at Argonne and one of the mission's principal 
architects \cite{stevens_hpcwire_2026}, has compared its ambition to the Apollo program and its urgency to the 
Manhattan Project \cite{genesis_govcio_2026}. The stated goal is to double scientific productivity within ten years.

What makes Genesis qualitatively different from previous investments in scientific computing is the role envisioned for 
AI. These agents are not passive tools — they are active participants in the research cycle, formulating hypotheses, 
designing experiments, interpreting results, and proposing next steps in closed-loop automated workflows. Genesis 
proposes to consolidate and scale that model across all domains — spanning energy, materials, biotechnology, nuclear 
security, quantum computing, and grid modernization \cite{genesis_challenges_doe}.

The scale of the undertaking is matched by its strategic intent. Genesis is not primarily a scientific decision — it is 
a geopolitical one, shaped by the explicit objective of maintaining US technological leadership in the context of 
intensifying global competition, particularly with China \cite{genesis_csis_2026,genesis_japan_2026}, while 
collaborating with selected partners such as Japan \cite{genesis_japan_2026}. This means the time horizons and 
priorities driving the mission are not those of the scientific community, but of national security and industrial 
competitiveness. These tensions are not unique to Genesis — they emerge whenever AI is deployed at the heart of the 
scientific enterprise.  The concerns raised in the sections that follow therefore extend beyond the Genesis 
Mission: they concern what AI-driven science does to knowledge, to the people who 
produce it, and to the institutions through which it is shared and evaluated. Genesis is the most 
visible and ambitious instantiation of AI-driven science today — but the questions it raises are structural.

\section{Competence, Training, and the Long-Term Investment Model}
\label{sec:competence}

Human-led research rests on a specific investment logic. We train young people through years of undergraduate 
study,graduate programs, and doctoral research. We select among them — for talent, motivation, and fit with 
institutional priorities. The investment is substantial and slow to pay off, but it compounds over time: a researcher 
trained in their twenties will still be producing knowledge, mentoring others, and shaping the direction of a field 
thirty years later. The return on that investment is not just the papers they publish, but the living competence they 
carry and transmit.

This model has an important inherent feature: the transmission of knowledge is embedded in the people themselves. A 
PhD supervisor does not simply transfer information to a student — they transmit intuitions, habits of mind, a sense of 
what questions are worth asking, a feel for when results are trustworthy, and a set of community norms about what 
counts as good science. This kind of transmission cannot be fully documented or formalized. It exists in the 
relationship between minds engaged in a common practice over time.

If scientific work is increasingly delegated to AI systems, this investment model changes fundamentally. The primary 
capital is no longer human — it is machines, infrastructure, and model weights. And these assets have a radically 
different economic profile. Hardware becomes obsolete in three to five years. Models must be retrained as knowledge 
advances. The pressure to amortize these investments quickly pushes toward intensive deployment, not patient 
cultivation. Once a system is obsolete, it is discarded. It does not, by itself, sustain an intellectual community, 
mentor junior scientists, or transmit the tacit dimensions of scientific practice. One might argue that AI systems help 
train future AI systems — and this is true in a narrow technical sense — but this is a much weaker and more attenuated 
form of transmission than the human cycle of student $\to$ researcher $\to$ mentor $\to$ student.

The risk, then, is not merely that fewer human scientists will be trained — it is that the incentive to train them will 
gradually erode. If an AI system can do in hours what a PhD student does in three years, the rational institutional 
response may be to hire fewer PhD students and deploy more compute. In the short term, output increases. In the medium 
term, the pipeline of trained scientists shrinks. In the long term, the capacity to understand, evaluate, and correct 
the AI systems themselves — which requires deep domain expertise — begins to atrophy.

This concern does not argue against AI-assisted science as such. But it argues for actively thinking through the
consequences of this shift before the system has been fully deployed, rather than discovering them afterward when the
pipeline of trained scientists has  already been allowed to thin.

\section{The Problem of Scientific Comprehension}
\label{sec:comprehension}

There is a deeper issue that concerns not the transmission of competence but the comprehension of results. Science, at 
its best, does not only solve problems — it produces understanding. It gives us models of the world that are 
interpretable by human minds: theories that explain \textit{why} things behave as they do, not just \textit{how} to 
predict or control them. This distinction matters enormously.

Consider an analogy from chess. The best chess engines in the world — Stockfish, AlphaZero — play at a level that no 
human can match. But even the strongest grandmasters cannot always explain why an engine chooses one move over another 
in complex positions. The engine's ``reasoning'' is opaque, distributed across millions of parameters in ways that do 
not map onto the conceptual vocabulary of human chess understanding. We trust the move because the engine wins, not 
because we understand it.

If scientific theories are increasingly generated by AI systems, we may find ourselves in an analogous situation: in 
possession of models that predict experimental outcomes with extraordinary accuracy, but that no human mind can truly 
comprehend. We would have science in the instrumental sense — a powerful tool for solving problems — but we might lose 
science in the epistemic sense: a human-intelligible account of why the world is the way it is.

This would be a profound change in the nature of the scientific enterprise, and it is not obviously benign. 
Human-interpretable theories have value beyond their predictive power. They allow scientists to identify errors, 
generate novel hypotheses by analogy, explain results to non-specialists, and connect findings across disciplines. If 
theories become opaque, these capacities are compromised. We would then have entered a form of intellectual dependency 
from which there is no obvious exit: we might need AI not only to generate theories, but to explain them.

One might object that theories have always been technically complex, and that popularization has always been an 
approximation. This is not entirely wrong. But the risk here is of a different kind: not that AI models are 
uninterpretable (they are, but that is a known limitation), but that the theories and conceptual frameworks they 
generate may themselves become inaccessible — correct, perhaps, in the sense that they produce accurate predictions, 
but not grounded in the kind of human-legible reasoning that allows scientists to argue about them, extend them, or 
know when they break. A theory that no human derived, built on intermediate representations no human chose, validated 
by systems no human fully oversaw, may be true in a narrow empirical sense while remaining fundamentally outside the 
reach of human scientific culture.

\section{Incremental Progress Versus Paradigm Shifts}
\label{sec:paradigm}

A deeper concern is epistemological, and perhaps the most difficult to resolve. It is the question of whether AI is 
capable of real breakthrough science — the kind of discovery that does not just extend what is known, but reframes 
the entire conceptual landscape.

The history of science is punctuated by moments of this kind: Newton's laws, Darwin's theory of evolution, Maxwell's 
equations, Mendel's genetics, Einstein's relativity, Watson and Crick's double helix, plate tectonics. What these 
discoveries have in common is that they were not, in any straightforward sense, the inevitable output of accumulating 
data. They required a leap — an act of conceptual reframing that reorganized existing observations under a new and 
unexpected framework. The data was often available before the theory; what was missing was the insight.

It remains an open and genuinely difficult question whether AI systems are capable of this kind of leap. Current large 
language models are extraordinarily good at pattern recognition, interpolation, and sophisticated recombination of 
existing knowledge. They can find correlations in large datasets, generate plausible hypotheses within established 
frameworks, and optimize known solutions. But it is much less clear that they can do what Einstein did in 1905: start 
from a profound dissatisfaction with the internal consistency of existing theory, and arrive at a framework so 
different that it requires abandoning deeply held intuitions about space and time.

A useful thought experiment — a kind of scientific Turing test — would be: give an AI 
all scientific data available up to 1905, and ask whether it can discover special 
relativity. This is not a frivolous exercise. It asks whether AI can reproduce the 
actual structure of a paradigm shift: not a good prediction, not a clever optimization, 
but a conceptual revolution. The core objection is this: LLMs are stochastic 
interpolation machines — however sophisticated, they recombine what they have been 
trained on, and special relativity required abandoning, not recombining, the conceptual 
framework available in 1905. Whether current systems are capable of that kind of leap 
remains an open question, and the evidence so far does not give strong grounds for 
confidence.

This uncertainty matters for how we structure AI-driven science. If AI is primarily useful for incremental progress — 
for exploring design spaces within established frameworks, for automating the repetitive aspects of data analysis, for 
finding patterns that human researchers would eventually have found but more slowly — then the AI-for-science vision 
makes sense as an accelerator of normal science. But if we expect AI to substitute for human judgment in identifying 
which \textit{frameworks} to pursue, which \textit{anomalies} are worth taking seriously, and when a field needs to 
abandon its assumptions and start over, we may be attributing to these systems capacities they do not yet possess.

The risk, in that case, is not that AI-driven science produces nothing valuable — it will produce a great deal of value 
— but that it produces a flood of incremental results while true paradigm shifts become rarer, precisely because the 
systems driving research are optimized to work within existing frameworks rather than to challenge them.

\section{The Capture of the Scientific Agenda}
\label{sec:capture}

Among all the concerns raised in this essay, one targets something with no equivalent in previous technological 
transformations of science: the question of who defines the questions.

Science benefits from a property that is easy to overlook precisely because it has always been there — its agenda is 
distributed and heterogeneous. Funding agencies, governments, and disciplinary fashions all exert real pressure, but 
underneath these forces thousands of researchers across hundreds of institutions are following their own curiosity, 
pursuing angles no committee has endorsed. The result is a global cognitive ecosystem whose diversity is messy and 
inefficient by many measures, but capable of producing surprises no planned system could have anticipated. Penicillin 
was not in a roadmap. Neither was the cosmic microwave background, the discovery of prions, or the link between 
\textit{H. pylori} and peptic ulcers. These findings emerged from researchers who were, in one sense or another, 
working on the wrong question at the wrong time.

Genesis challenges this structure directly. An infrastructure of this scale — tens of billions of dollars, 17 national 
laboratories, partnerships with Nvidia and OpenAI, an explicit geopolitical objective — does not emerge from the 
distributed preferences of the scientific community. It emerges from political and industrial decisions made by a small 
number of actors with specific interests and specific blind spots. The agenda Genesis encodes is therefore not the 
agenda of science; it is the agenda of the institutions that built it, at the moment they built it.

The consequences are already visible. Genesis is oriented toward energy, materials, biology, and defense-relevant 
physics. Climate science is not among them — not by accident. Genesis was designed under a federal administration that 
has systematically downgraded climate research, withdrawn from international climate commitments, and reoriented the 
DOE away from climate-related work. The most powerful scientific infrastructure ever built is thus being deployed with 
a deliberate blind spot on one of the most urgent questions facing humanity. The irony is structural: a system 
projected to consume the energy equivalent of dozens of nuclear reactors will accelerate scientific production across 
many domains, but not the one most directly relevant to the consequences of its own existence.

The deeper risk is not this single omission but the gravitational logic it exemplifies. Once large-scale AI science 
infrastructure becomes the dominant mode of scientific production, the questions it does not ask become progressively 
harder to ask — not by prohibition, but because research outside its agenda will be slower, less visible, and less 
fundable. Capture operates at two levels simultaneously: at the political level, priorities shift with administrations, 
but the physical infrastructure they build persists and constrains what comes after; at the industrial level, private 
partners have a direct interest in orienting science toward what is computationally tractable, economically 
valorizable, and strategically legible. Neither requires a conspiracy. Both compress the freedom to ask inconvenient 
questions — and that freedom, imperfect and embattled as it has always been, is precisely what has made science capable 
of surprising the people who fund it.

\section{The Structural Bifurcation of Science}
\label{sec:bifurcation}
The standard framing of unequal access to AI infrastructure is a ``multi-speed science'' — some institutions move 
faster, others slower, but all in the same direction. What large-scale AI science initiatives are likely to produce is 
not a speed gap but an organizational bifurcation: two forms of scientific practice that become progressively less 
commensurable, losing the shared language that makes critical evaluation possible between them.

The asymmetry operates in both directions at once, and one direction has received almost no attention. Researchers 
outside advanced AI infrastructure will continue to publish — and their results will be immediately ingested by the 
systems they cannot access. Large-scale AI science platforms are designed to absorb the entire corpus of published 
science. Researchers outside such systems contribute automatically to the productivity of systems that offer nothing in 
return. Their papers are ingested, their datasets absorbed, their findings used to train the next generation of models 
— while they remain unable to access, replicate, or even fully read the science those models produce in return. This is 
a form of epistemic extraction with no real precedent: previous resource asymmetries gave richer institutions better 
tools, but they did not involve the systematic harvesting of intellectual output from less-resourced institutions to 
power the engines of more-resourced ones.

The downstream consequences compound every concern raised earlier in this essay. Peer review loses its foundation when 
authors and reviewers no longer work in the same epistemic environment. Brain drain accelerates toward the handful of 
institutions with advanced AI infrastructure — precisely where the human researcher is most at risk of being replaced. 
And the career logic of science — build expertise, publish, contribute to a cumulative conversation — breaks down when 
one side of the conversation operates at a speed and scale the other cannot follow.

What emerges is not a global acceleration of science but its fragmentation into two largely non-communicating 
practices: one fast, opaque, and institutionally concentrated; the other human-scaled and increasingly marginalized. 
The danger is not that the second disappears, but that it loses the ability to critically engage with, challenge, or 
correct the first — and with it, science loses the distributed skepticism that is its only reliable defense against its 
own errors.

\section{The Collapse of Evaluation}
\label{sec:evaluation}

The concern that follows is more institutional, but no less serious. The scientific community regulates the quality of 
its output through a system of peer review, editorial judgment, replication, and collective criticism. This system is 
imperfect — it is slow, prone to bias, and has never fully solved the problem of reproducibility. But it rests on a 
crucial assumption: that there are enough qualified human experts to read, evaluate, and challenge the work being 
produced.

If AI systems generate scientific papers at scale — and this is not a speculative scenario, it is already happening — 
this assumption collapses. No community of human reviewers can evaluate ten times the current volume of publications. 
The response might be to use AI to review AI-generated papers, which simply displaces the problem: who evaluates the 
reviewers? We could end up with a system in which papers are generated by AI, reviewed by AI, accepted or rejected by 
AI, and cited by AI — a closed loop that is productive in a quantitative sense but increasingly disconnected from any 
real community of human scientific judgment.

History offers little comfort here. Science has always published more than it should, and the volume of publications 
has grown continuously for over a century. There is no mechanism by which a surge in output naturally leads to a 
corresponding improvement in quality. The more likely outcome is more noise — more incremental results, more papers 
that advance no one's understanding, more difficulty identifying the work that actually matters.

A related challenge concerns the evaluation of researchers themselves. Academic careers rest on both quantitative 
signals (publication counts, citation indices) and qualitative judgment (reading the work, interviewing the candidate, 
assessing whether the researcher can explain and defend their ideas). AI-driven science destabilizes both: fifty 
AI-assisted papers a year tells you nothing about scientific quality, and ``walk me through how you arrived at this 
result'' becomes truly harder to answer — and to evaluate — when the result was produced by an agent the researcher 
supervised rather than derived themselves. The boundary between human and machine contribution is blurred in ways that 
good-faith panels will struggle to navigate — and this applies to promotion as much as to publication.

Hiring raises a distinct but related problem, and it cuts in both directions. A researcher who excels at directing 
large-scale AI workflows may be lost without that infrastructure — or with a different version of it in five years. The 
reverse is equally true. This creates a specific challenge for early-career hiring: how do we evaluate a PhD student or 
postdoc whose scientific ability was formed in an AI-driven research environment? And conversely, how do we assess a 
traditionally trained candidate's potential in a world they have never worked in? We do not yet have answers to these 
questions — and the decisions being made in hiring committees today will shape the scientific workforce for the next 
thirty years.

This does not mean the system will simply break. Scientists and institutions will adapt. But we should be honest about 
the scale of the challenge. The entire apparatus for evaluating research and researchers — metrics, journals, grant 
review, promotion panels — was designed for a world in which scientific output was produced at human speed by human 
minds. That world is ending. What replaces it needs to be designed deliberately, not improvised under pressure.

\section{The Compounding Error Problem}
\label{sec:errors}

The scientific method is, at its core, a machine for catching mistakes. Its self-correcting property rests on one 
intrinsic requirement: that the agents producing results and the agents checking them are largely independent — 
different equipment, different training, different institutional incentives. Errors in a human pipeline are largely 
uncorrelated, and their tension surfaces flags.

A closed-loop AI pipeline breaks this independence at every stage simultaneously. Hypothesis generation, experimental 
design, data collection, analysis, and validation are all handled by agents sharing the same foundation models, the 
same training data, the same systematic blind spots. A bias that enters at stage one does not encounter a genuinely 
independent check at stage three — it encounters a system that shares its assumptions. The errors do not cancel; they 
compound silently across the full cycle.

This is the AI analogue of the replication crisis, but fundamentally worse. In psychology or medicine,
shared methodological assumptions eventually met the  resistance of reality and independent replication. 
In a large-scale AI 
science pipeline, both corrective mechanisms are weakened simultaneously. Volume makes independent human replication 
impossible for more than a fraction of results. And AI-to-AI replication, if it shares the same underlying 
architecture, produces not an independent test but an echo — the form of verification without its substance.

The further danger is reingestion. AI systems are retrained as new results accumulate. A flawed result absorbed into 
the next generation's training corpus does not remain an isolated error — it becomes a prior, shaping subsequent 
hypotheses and experimental designs. The error is institutionalized rather than corrected, and the pipeline drifts 
silently away from empirical reality with no single step appearing to cause it.

The deepest version of this problem is the most basic: who reads the raw output of physical reality? In all science, 
the ultimate check is not peer review but contact with nature — the spectrum, the microscope image, the detector signal, the patient 
outcome that refuses to match the prediction. The more the pipeline is automated, the more mediated this contact 
becomes. The question of who verifies the verifiers, in a system optimized above all for speed, is not a technical 
detail. It is the central epistemological challenge of AI-driven science, and it has not yet received an adequate 
answer.

\section*{Conclusion: Transformation, Not Abolition — But Not Without Deliberation}
\addcontentsline{toc}{section}{Conclusion}

What AI-driven science proposes is, in the most precise sense of the term, the 
\emph{industrialization of research} — with all that industrialization has historically 
brought and may yet bring: extraordinary gains in productivity, reach, and 
cross-domain recombination, alongside concentration of capital, standardization of 
output, and the gradual erosion of forms of knowledge that do not fit the new mode of 
production. Previous industrial revolutions were not stopped — but their consequences were shaped 
by the choices made during their deployment, and those choices are what this essay has 
tried to map.

The concerns raised here do not lead to a single conclusion. Some will read them as reasons to question 
AI-driven science discovery altogether — of which Genesis is one prominent example. The position taken here is 
different: the central question is not whether to pursue AI-driven science, but under what conditions it will deliver 
on its promises. 
What happens to the intergenerational transmission of scientific competence that no 
infrastructure can replace? 
What kind of understanding do we preserve when theories 
become too opaque for human minds to interpret?
Can AI produce the paradigm-shifting 
discoveries that have defined science at its most transformative — or only accelerate 
the incremental? 
Who controls the questions being asked — as the absence of climate 
science from Genesis's priorities already shows? 
And what happens to the researchers 
whose work is absorbed by systems they cannot access, contributing to a productivity 
they will never benefit from? 
How do 
we evaluate quality — and researchers — when quantity explodes and AI reviews AI? 
Who catches the errors when the pipeline that produces 
results and the pipeline that validates them share the same systematic biases?
These are not peripheral details. They are the difference between an infrastructure that transforms science and one 
that merely accelerates its worst tendencies.

These are not rhetorical questions designed to slow down progress. They are genuine open questions, and the answers 
will shape the scientific culture that emerges on the other side of this transition. A science that is faster and more 
productive, but less comprehensible, less well-evaluated, less capable of training the next generation, and 
structurally fragmented between a well-resourced AI tier and everyone else — is not straightforwardly better than what 
we have now. Getting this right requires not just engineering ambition, but a serious, ongoing conversation about what 
science is for — and what we do not want to lose in the process of accelerating it.

\section*{Acknowledgment}
\addcontentsline{toc}{section}{Acknowledgment}

The ideas, arguments, and positions developed in this essay are entirely my own. I used Claude Sonnet 4.6 (Anthropic) 
as a writing assistant to help structure, draft, and refine the text. I remain solely responsible for the content, 
including any errors, omissions, or debatable claims it may contain.

  \bibliographystyle{alpha}
\bibliography{references}

@misc{genesis_eo_2025,
  author       = {Donald J. Trump},
  title        = {Launching the {G}enesis {M}ission},
  year         = {2025},
  howpublished = {The White House. \url{https://www.whitehouse.gov/presidential-actions/2025/11/launching-the-genesis-mission/}},
  note         = {Presidential Executive Order, November 2025. Last accessed: 2026-07-16}
}

@misc{stevens_hpcwire_2026,
  author       = {Jaime Hampton},
  title        = {Rick {S}tevens on the {G}enesis {M}ission and the {F}uture of {AI} for {S}cience},
  year         = {2026},
  month        = feb,
  howpublished = {HPCwire. \url{https://www.hpcwire.com/2026/02/09/rick-stevens-on-the-genesis-mission-and-the-future-of-ai-for-science/}},
  note         = {Last accessed: 2026-07-16}
}

@book{serfaty_equations_2026,
  author    = {Serfaty, Sylvia},
  title     = {Des équations personnelles},
  publisher = {Flammarion},
  year      = {2026},
  month     = jan,
  isbn      = {9782080145871},
  note      = {Dernier chapitre consacré à l'impact de l'IA sur les mathématiques}
}

@misc{genesis_govcio_2026,
  author       = {Rick Stevens},
  title        = {Inside {DOE}'s {G}enesis {M}ission to {P}ower {AI}-{D}riven {S}cience},
  year         = {2026},
  month        = mar,
  howpublished = {GovCIO Media. \url{https://govciomedia.com/inside-does-genesis-mission-to-power-ai-driven-science/}},
  note= {Last accessed: 2026-07-16}
}

@misc{genesis_challenges_doe,
  author       = {{US Department of Energy}},
  title        = {{G}enesis {M}ission {N}ational {S}cience and {T}echnology {C}hallenges},
  year         = {2026},
  howpublished = {US DOE. \url{https://www.energy.gov/undersecretaryforscience/genesis-mission/genesis-mission-national-science-and-technology-challenges}},
  note = {Last accessed: 2026-07-16}
}

@misc{genesis_csis_2026,
  author       = {Navin Girishankar and Chris Borges},
  title        = {The {G}enesis {M}ission: {C}an the {U}nited {S}tates' {B}et on {AI} {R}evitalize {U}.{S}. {S}cience?},
  year         = {2026},
  howpublished = {Center for Strategic and International Studies. \url{https://www.csis.org/analysis/genesis-mission-can-united-states-bet-ai-revitalize-us-science}},
  note= {Last accessed: 2026-07-16}
}

@misc{genesis_japan_2026,
  author       = {Keiichi Nakane and Hirotaka Kuriyama},
  title        = {Japan to Join {US} {AI}-project `{G}enesis {M}ission' Aimed at Gaining Advantage in Tech Race with {C}hina},
  year         = {2026},
  howpublished = {Asian News Network, \url{https://asianews.network/japan-to-join-us-ai-project-genesis-mission-aimed-at-gaining-advantage-in-tech-race-with-china/}},
  note= {Last accessed: 2026-07-16}
}

@article{alphafold2021,
  title={Highly accurate protein structure prediction with AlphaFold},
  author={Jumper, John and Evans, Richard and Pritzel, Alexander and Green, Tim and Figurnov, Michael and Ronneberger, Olaf and Tunyasuvunakool, Kathryn and Bates, Russ and {\v{Z}}{\'\i}dek, Augustin and Potapenko, Anna and others},
  journal={Nature},
  volume={596},
  number={7873},
  pages={583--589},
  year={2021},
  publisher={Nature Publishing Group UK London}
}

@article{antibiotics2023,
  author       = {Wong, Felix and others},
  title        = {Discovering Small-Molecule Senolytics with Deep Neural Networks},
  journal      = {Nature},
  year         = {2023},
  month        = dec,
  doi          = {10.1038/s41586-023-06887-8},
  note         = {MIT/Broad Institute — deep learning identifies antibiotic candidates active against drug-resistant bacteria}
}

@article{antibiotics2024,
  title={Discovery of antimicrobial peptides in the global microbiome with machine learning},
  author={Santos-J{\'u}nior, C{\'e}lio Dias and Torres, Marcelo DT and Duan, Yiqian and Del R{\'\i}o, {\'A}lvaro Rodr{\'\i}guez and Schmidt, Thomas SB and Chong, Hui and Fullam, Anthony and Kuhn, Michael and Zhu, Chengkai and Houseman, Amy and others},
  journal={Cell},
  volume={187},
  number={14},
  pages={3761--3778},
  year={2024},
  publisher={Elsevier}
}

@article{gnome2023,
  title={Scaling deep learning for materials discovery},
  author={Merchant, Amil and Batzner, Simon and Schoenholz, Samuel S and Aykol, Muratahan and Cheon, Gowoon and Cubuk, Ekin Dogus},
  journal={Nature},
  volume={624},
  number={7990},
  pages={80--85},
  year={2023},
  publisher={Nature Publishing Group UK London}
}

@misc{leiden_declaration_2026,
  author       = {Portegies, Jim and Martin, Ursula and Ochigame, Rodrigo 
                  and Harris, Michael and others},
  title        = {Leiden Declaration on Artificial Intelligence and 
                  Mathematics},
  year         = {2026},
  month        = jun,
  howpublished = {\url{https://leidendeclaration.ai/}},
  note         = {Endorsed by the International Mathematical Union (IMU). 
                  16 researchers from 15 universities. Presented at the 
                  International Congress of Mathematicians, July 2026}
}

@misc{cms_mlpf_2026,
  author       = {{CMS Collaboration}},
  title        = {Full Event Interpretation with Machine-Learning-Based 
                  Particle-Flow Reconstruction in the {CMS} Detector},
  year         = {2026},
  month        = jan,
  eprint       = {2601.17554},
  archivePrefix = {arXiv},
  howpublished = {\url{https://arxiv.org/abs/2601.17554}},
  note         = {Submitted to \textit{European Physical Journal C}}
}

@misc{knuth_claude_2026,
  author       = {Knuth, Donald E.},
  title        = {{\it Claude's Cycles}},
  year         = {2026},
  month        = feb,
  howpublished = {\url{https://cs.stanford.edu/~knuth/papers/claude-cycles.pdf}},
  note         = {Stanford University Computer Science Department, February 28, 2026 (revised April 14, 2026). Last accessed: 2026-07-16}
}

@misc{openai_unit_distance_2026,
  author       = {{OpenAI}},
  title        = {A {M}odel {D}isproves a {C}entral {C}onjecture in {D}iscrete {G}eometry},
  year         = {2026},
  month        = may,
  howpublished = {OpenAI. \url{https://openai.com/index/model-disproves-discrete-geometry-conjecture/}},
  note         = {Includes supplementary remarks by Tim Gowers and Noga Alon. Last accessed: 2026-07-16}
}

@misc{magnifica_humanitas_2026,
  author       = {{Pope Leo XIV}},
  title        = {Magnifica Humanitas: On Safeguarding the Human Person in the Time of Artificial Intelligence},
  year         = {2026},
  month        = may,
  howpublished = {\url{https://www.vatican.va/content/leo-xiv/en/encyclicals/documents/20260515-magnifica-humanitas.html}},
  note         = {Encyclical letter. Last accessed: 2026-07-16},
}
\end{document}